# Evolutionary model for energy trading in community microgrids using Hawk-Dove strategies

Viorica Rozina Chifu, Tudor Cioara, Cristina Bianca Pop, Ionut Anghel

Technical University of Cluj-Napoca, Department of Computer Science, Romana

This paper proposes a decentralized model of energy cooperation between microgrids, in which decisions are made locally, at the level of the microgrid community. Each microgrid is modeled as an autonomous agent that adopts a Hawk or Dove strategy, depending on the level of energy stored in the battery and its role in the energy trading process. The interactions between selling and buying microgrids are modeled through an evolutionary algorithm. An individual in the algorithm population is represented as an energy trading matrix that encodes the amounts of energy traded between the selling and buying microgrids. The population evolution is achieved by recombination and mutation operators. Recombination uses a specialized operator for matrix structures, and mutation is applied to the matrix elements according to a Gaussian distribution. The evaluation of an individual is made with a multi-criteria fitness function that considers the seller profit, the degree of energy stability at the community level, penalties for energy imbalance at the community level and for the degradation of microgrids batteries. The method was tested on a simulated scenario with 100 microgrids, each with its own selling and buying thresholds, to reflect a realistic environment with variable storage characteristics of microgrids batteries. By applying the algorithm on this scenario, 95 out of the 100 microgrids reached a stable energy state. This result confirms the effectiveness of the proposed model in achieving energy balance both at the individual level, for each microgrid, and at the level of the entire community.

## 1. Introduction

In many regions with weak or non-existent electrical infrastructure, traditional grids often face severe intermittency, high grid losses, and high costs for expanding the national grid. In this context, local community microgrids have become a promising solution for electrifying isolated communities, facilitating local energy production and consumption, especially from renewable sources such as solar or wind energy. Unlike traditional grids that fall into the category of centralized systems, these microgrids operate most often without a central distribution system operator (DSO). Each microgrid has operational autonomy and stores energy locally, usually in batteries, managing production, consumption and possible energy exchanges with other microgrids. The decentralization at the community level brings significant challenges, including maintaining energy balance at the community level, efficient use of storage capacities, and preventing accelerated battery degradation. In the absence of a distribution system operator (DSO), voluntary collaboration between microgrids becomes essential for the efficient management of energy surpluses and deficits. Ideally, microgrids with surplus should be able to supply energy to those with deficits, thus contributing to the energy stability of the entire community. However, these interactions are influenced by a number of factors, including economic objectives, such as maximizing profit or conserving resources, but also physical constraints, such as line capacity or the risk of battery degradation. In this context, peer-to-peer (P2P) energy trading plays a key role because it allows microgrids from energy communities to share their energy surplus with neighboring ones facing a deficit [3]. This improves security of supply without the need for a centralized control entity. Furthermore, by localizing energy exchanges, P2P trading reduces the need for long-distance transport, which leads to lower costs and energy losses [4].

This paper proposes a decentralized cooperative model for energy trading between microgrids in a community microgrids, inspired by evolutionary game theory. The seller microgrids adopt Hawk or Dove behavior, depending on the energy level in the battery and the overall energy stability at the community microgrids. Hawk seller microgrids initiate energy sales when the storage level of battery exceeds an upper threshold, to maximize their payoff, accepting an increased risk of battery degradation. In contrast, Dove microgrids adopt a more cautious sales policy, trading only above a lower threshold. Microgrids acting as buyers adopt the same type of behavior: any microgrid in energy deficit tries to purchase energy to reach equilibrium, without applying a specific trading strategy, such as Hawk or Dove. The optimization of energy trading decisions is achieved by a genetic algorithm. Each individual is represented by an energy matrix, where each element encodes the quantity of energy trading between a selling and a buying microgrid. The matrix respects two constraints: no transactions are allowed between microgrids of the same type (i.e. seller microgrids or buyer microgrids), and self-trading is excluded (a microgrid cannot transact with itself). The quality of each individual is evaluated through a multi-criteria fitness function, which integrates: (i) the total profit obtained by the selling microgrids; (ii) the number of microgrids that reach an energy balance state at the end of the transaction; and (iii) an aggregate penalty, computed according to the deviations from the stable state of the community microgrids, the frequency of use of the Hawk strategy by the selling microgrids, the estimated degree of battery degradation and the level of overloading of distribution lines during the transaction process. Recombination is done by applying a recombination operator specific to matrix operations, and mutations are applied to individual, according to a Gaussian distribution. The method was tested on a simulated scenario involving 100 microgrids, each with its own energy buy and sale thresholds. This setup was designed to simulate a realistic environment, in which a community of heterogeneous microgrids interact, each with its own battery storage characteristics.

Unlike existing approaches in the research literature, which assume either centralized control or the application of static trading rules lacking adaptability to the local context, the model proposed introduces a completely decentralized architecture, in which energy balancing decisions are taken autonomously by each microgrid. The trading behaviors are modeled using evolutionary game theory, which allows the description of the cooperation and competition relations between the selling microgrids in the process of supplying energy to the buying microgrids, with the aim of maintaining the stability of the community microgrids. In addition, the integration of an adaptive genetic algorithm allows the identification of trading configurations that provide the best results at the community level, based on a multi-criteria fitness function.

The structure of the paper is as follows: Section 2 reviews relevant works in the field. Section 3 details the proposed evolutionary cooperative model for energy trading in community microgrids. Section 4 is dedicated to the performance evaluation of this model, while Section 5 discusses aspects regarding the stability of the genetic algorithm used, as well as the balance between exploration and exploitation. The paper concludes with conclusions and future work directions.

## 2. Related Work

Many state-of-the-art approaches use game theory and deep reinforcement learning as techniques for P2P energy trading between energy communities or interconnected microgrids [14].

[8] uses Nash equilibrium to define an incentive mechanism facilitating the P2P energy trading between interconnected microgrids, with the aim of reducing costs. The proposed approach assumes that microgrids are independent and rational without being involved in an alliance or affected by the external environment. The objective of each microgrid is to minimize the total costs. [9] proposes a coalitional game theory-based method for local power exchange among networked microgrids. The goal is to enable nearby microgrids to trade energy such that their energy requirements are met and the individual utility is increased. The method uses auction theory to compute the maximum utility for each coalition and determine the trading order. Afterwards, the optimal coalition partition is identified using coalition merging and splitting. Nash bargaining is used in [16] to model multi-microgrid multi-energy and communication trading and to ensure an equitable distribution of trading gains. [17] integrates Nash bargaining in a green electricity trading mechanism for interconnected microgrids to determine the trading capacity. The allocation of green electricity values between buyers and sellers is done based on an environmental factor computed based on carbon emission and energy composition. [18] models the trading process among micro-grids as a multi-leader multi-follower Stackelberg game. [15] introduced the energy trading consistency concept to avoid unreasonable trading actions in P2P energy trading between interconnected multi-energy microgrids. The authors formulate the P2P energy trading problem as a partially observable Markov decision process and apply deep reinforcement learning to solve it, ensuring that trading actions—such as buying or selling energy—stay within the constraints of transmission line capacity. [7] uses deep reinforcement learning to optimize P2P energy trading among multiple industrial, residential and commercial microgrids. The goal is to maximize trading profits while ensuring that microgrids trade energy only if the local energy demands are met and the remaining energy of a microgrid does not exceed the maximum capacity of the battery. In this approach, trading decision-making is modeled as a Markov game.

[14] uses a multi-agent deep reinforcement learning method to solve the P2P energy trading problem within interconnected heterogeneous multi-energy microgrids residential, commercial and industrial MEMGs. The method combines the multi-agent actor-critic algorithm with the twin delayed deep deterministic policy gradient algorithm. [5] address the problem of P2P trading between energy communities by using a bipartite graph as abstraction for the transaction process and a recurrent Hungarian algorithm for matching energy buyers and sellers. The matching process occurs after the intra-community P2P markets are cleared and the information about the energy communities having an energy excess/deficit is obtained. [12] proposes a computational efficient method for cross-microgrid P2P energy trading which uses clustering algorithms for grouping microgrids in coalitions based on their profiles (prosumers' socio-demographics, amounts of power the microgrid is intending to purchase or sell). The energy trading is initiated at the level of each microgrid and if there are internal trading requests that cannot be satisfied, microgrid coalitions are created

and energy trading continues between the prosumers, part of the same coalition. [6] proposes a decentralized bi-level energy trading framework that optimizes external interactions between microgrids and internal scheduling within microgrids. Principal component analysis is used for optimal trading and scheduling. The objective of a microgrid is to minimize the total operation and discomfort costs. The P2P trading aims to satisfy a set of balance constraints referring to the active and reactive power flow. [10] proposes a priority matching P2P trading mechanism between interconnected microgrids. Scheduling results and quotations are used to compute the priority indices reflecting the bargaining intent of each microgrid. Microgrids, part of matching pairs, are allowed to change their quotation several times, based on the pair's supply-demand ratio. [11] proposes a distributionally robust optimization strategy for P2P energy trading between multiple microgrids to handle renewable energy uncertainty using fuzzy sets, and a privacy-preserving distributed algorithm. With the aim of optimizing energy exchanges between geolocated microgrids, [13] proposes a coalition formation algorithm that reduces energy loss, energy costs, and energy storage systems degradation costs. Buyer microgrids that have the highest demand initiate coalition formation by selecting seller microgrids considering criteria such as geographic proximity, transmission costs and economic benefits.

Our approach differs from existing approaches in the literature by introducing a complete decentralized framework for energy trading between microgrids, based on evolutionary game theory and optimized by adaptive genetic algorithms. While previous works rely either on centralized coordination mechanisms [8], [9], [16], or on static or rigid rule-controlled models [18], [15], [14], our model allows microgrids to make autonomous decisions depending on the local context and the dynamics of the energy system. Modeling hawk-dove behaviors allows capturing competitive and cooperative relationships between microgrids, with the aim of ensuring the stability of the energy community, in a more realistic and flexible way than the classical Nash or Stackelberg bargaining models. Furthermore, integrating battery degradation into the decision-making process introduces a practical dimension often neglected in previous work, ensuring the sustainability of long-term transactions. The adaptive genetic algorithm identifies the most efficient trading configurations at the community level, using a multi-criteria fitness function that not only optimizes energy usage, but also discourages line overloads and aggressive seller behaviors, thus ensuring energy balance across the entire community and promoting a fair distribution of resources.

# 3. The evolutionary cooperation model for energy trading in community microgrids

We consider a community of interconnected microgrids formally defined as:

$$MGN = \{MG_i | \ i = \overline{1,..,n}\} \quad (1)$$

where $MG_i$ is a microgrid and *n* is the total number of microgrids in the community microgrids.

The microgrids are connected to each other through a virtual network, which allows energy trading between microgrids. The energy trading process is based on a theoretical model inspired by game theory that describes the interactions and behaviours between microgrids. Trading is peer-to-peer (P2P), which means that microgrids interact directly, making autonomous decisions in the process of energy negotiation and exchange.

In our approach, within the community microgrids, we consider that energy transactions are based on the energy available in the microgrid batteries. Each microgrid has as its main objective to maintain the battery energy level as close as possible to a predefined threshold, considered optimal for its operation, while contributing to the energy balance of the community of which it is part. The trading behaviour of a microgrid is governed by two thresholds:
- Buy Threshold (BT): This threshold represents the minimum level of energy stored in the battery, below which the microgrid adopts a buying behavior.
- Sell Threshold (ST): This threshold indicates the energy level stored in the battery above which the microgrid adopts a selling behavior.

The community microgrids is considered stable at time *t* if the level of energy stored in the batteries of each microgrid remains within the range defined by the BT and ST thresholds.

Mathematically, the stability state of the community is expressed as:

$$\forall t \in T, \forall i \in [1,n], \ MG_i(t) \ is \ stable \iff BT_i \leq E_{b_i}(t) \leq ST_i \quad (2)$$

where: $BT_i$ is the buying threshold of microgrid *i*, $ST_i$ is the selling threshold of microgrid *i*, $E_{b_i}(t)$ is the level of energy stored in the battery of microgrid *i*, after trading, at time *t*.

Within the community, selling microgrids involved in the trading process can adopt behaviours inspired by the Hawks and Doves, a game theory model used to describe competition for resources in distributed systems. Hawk seller microgrids initiate energy trading when the energy level in battery exceeds the ST sales threshold, with the aim of maximizing the traded quantity and the related profit. However, this behaviour comes with the risk of accelerating the battery degradation. In contrast, Dove microgrids prefer a more cautious approach, providing energy only when the battery level is above the BT threshold, even if they are already in a state of energy balance. In the proposed model, this classification into Hawk or Dove influences exclusively the selling behaviour. Regarding energy buying, the behaviour is the same, namely any microgrid that is in a state of energy deficit will try to buy energy to stabilize as quickly as possible.

In our approach, a microgrid MG$_i$ is formally defined at time *t* as follows:

$$MG_i(t) = <E_{b_i}(t), role_i, (BT_i, ST_i), S_i(t), B_i(t)>, E_i(t) \in [0, E_{b_i}^{max}] \quad (3)$$

where: $E_{b_i}^{max}$ is the maximum amount of energy that can be stored in the battery; $E_{b_i}(t)$ is the energy available in battery at time *t*; role$_i$ ∈ {Hawk, Dove, None} defines the strategy adopted by the microgrid in the trading process; BT$_i$ and ST$_i$ are buying and selling thresholds; S$_i$(t) is the amount of energy that the microgrid could sell at time *t*, determined by the energy surplus in the battery; B$_i$(t) is the amount of energy that the microgrid would be willing to buy at time *t*, determined by the energy deficit.

A microgrid can have only one trading intention at a time *t*, either to sell or to buy, or not to participate in the transaction. The amount of energy that a microgrid could sell at time *t*, respectively the amount of energy that a microgrid could buy at time *t* is formally defined as follows:

$$S_i(t) = \begin{cases} E_{b_i}(t), -ST_i, if\ E_{b_i}(t) > ST_i \vee E_{b_i}(t) > BT_i \\ otherwise \end{cases} \quad (4)$$

$$B_i(t) = \begin{cases} BT_i - E_{b_i}(t), if\ E_{b_i}(t) < BT_i \\ otherwise \end{cases} \quad (5)$$

The role of a microgrid $MG_i(t)$ determines its trading behaviour, depending on the current energy level:

$$strategy_i(t) = \begin{cases} Hawk\ seller, if\ E_{b_i}(t) > ST_i \wedge role_i = Hawks \\ Dove\ seller, if\ E_{b_i}(t) > BT_i \wedge role_i = Dove \\ Buyer, if\ E_{b_i}(t) < BT_i \wedge role_i = None \\ None \end{cases} \quad (6)$$

In our approach, the trading thresholds BT and ST are set per microgrid and determine the stability interval for the microgrid:

$$0 \leq BT_i \leq ST_i \leq E_{b_i}^{max} \quad (7)$$

where $E_{b_i}^{max}$ is the maximum energy that can be stored in the battery of the microgrid *i*.

In addition, the proposed model integrates a global transaction threshold, denoted $THV \in \mathbb{R}_+$, which sets the upper limit on the amount of energy that can be transferred in a single interaction between two microgrids. The introduction of this threshold reflects both real technical constraints (i.e. power line capacity, or local storage limitations) and algorithmic requirements related to the stability of the community microgrids. Limiting the volume per transaction prevents the occurrence of excessive energy transfers that could unbalance the community and, at the same time, promotes a fair distribution of energy within the community microgrids.

$$E_{b_i \to b_j}(t) \leq THV \quad (8)$$

In this context, to identify the optimal microgrid configurations between which energy transfers can take place to ensure the stability of community microgrids, respecting all network constraints, we propose the use of a genetic algorithm. Within this, each potential solution (i.e. an individual) is represented by a square matrix $EM \in \mathbb{R}_{nxn}$, where *n* is the total number of microgrids in the community.

$$indiv = EM(t) = \begin{bmatrix} 0 & E_{b_1 \to b_2}(t) & \ldots & E_{b_1 \to b_n}(t) \\ E_{b_2 \to b_1}(t) & 0 & \ldots & E_{b_2 \to b_n}(t) \\ \ldots & \ldots & \ldots & \ldots \\ E_{b_n \to b_1}(t) & E_{b_n \to b_1}(t) & \ldots & 0 \end{bmatrix} \quad (9)$$

Each element $E_{b_i \to b_j}(t)$ in this matrix expresses the intention to transfer energy, in units of kWh, from microgrid *MG$_i$* to *MG$_j$*, at time *t*. Although the matrix encodes all possible transactions between microgrid pairs, only a restricted subset of elements is activated, depending on the current energy state of the microgrids involved. More precisely, a transfer intention is considered valid only if MG$_i$ is in a state of energy surplus (i.e., $E_{b_i}(t)$ >ST$_i$) and can act as seller and *MG$_j$* in a state of deficit (i.e., $E_{b_j}(t)$ <BT$_j$) and can act as buyer, while respecting the condition , i ≠ j . Therefore, actual energy transactions take place exclusively between a microgrid seller and a microgrid buyer. Cases where both microgrids are in a stable energy state are excluded, as well as those where the transfer is proposed between microgrids of the same type (i.e. either two sellers or two buyers).

The proposed amounts for energy transfer are initially randomly generated in the initialization stage of the genetic algorithm and are subsequently refined through genetic operators of crossover and mutation, guided by the performance of each individual according to the fitness value.

It is important to note that the matrix EM(t) encodes only the transaction intentions and not the final volume of energy transferred. To obtain this volume, we construct the real transactions matrix, by applying a limiting function to each element of the EM(t) matrix:

$$E_{b_i \to b_j}(t)(t) = \min(E_{b_i}(t) - ST_i, BT_j - E_{b_j}(t), E_{b_i \to b_j}(t), THV) \quad (10)$$

where:
- $E_{b_i}(t) - ST_i$ represents the available energy surplus of the selling microgrid.
- $BT_j - E_{b_j}(t)$ reflects the energy deficit of the buying microgrid.
- $E_{b_i \to b_j}(t)$ is the transfer intention proposed by the genetic algorithm
- THV is the global transaction threshold, which imposes an upper limit on the amount of energy that can be transferred in a single interaction.

If both microgrids involved are in a stable state and no transactions are made between them, then: $E_{b_i \to b_j}(t) = 0$. Thus, the real transactions matrix is expressed based on these new computed values and encodes the volumes of energy actually transferred between valid pairs of microgrids at time *t*, considering all imposed energy constraints.

The fitness function used to evaluate each candidate individuals integrate the following components: (a) the total profit obtained by sellers from the transactions made; (b) the number of microgrids that reach an energy balance level,; (c) a global bonus granted when a significant proportion of the microgrids in the community reach a stable state following trading; and (d) a total penalty, which sanctions suboptimal individuals.

$$fitness(indiv) = \frac{\alpha * Payoff + \beta * noS + \gamma * B - P_{total}}{F_{max}} \quad (11)$$

where: *Payoff* represents the total profit obtained by the selling microgrids; *S* is the number of microgrids in a stable state after the end of the trading; *B* is the stability bonus, computed according to the proportion of stabilized microgrids; $P_{total}$ represents the total penalty, which integrates deviations from stability, strategy abuse, battery degradation and overloading level of distribution lines; $F_{max}$ is the theoretical maximum value of fitness, used to normalize the score in the interval [0, 1]; α, β, γ, are empirical weighting coefficients, which specify the influence of each component within the fitness function.

The total profit obtained by the selling microgrids following the trading process is computed by summing all quantities of energy transferred to buyers, each transaction being weighted according to its effect on the buyer's energy stabilization.

$$Payoff = \sum_{i=1}^{n} \sum_{\substack{i=1 \\ j \neq i}}^{n} E_{b_i \to b_j}(t) * \pi_{i \to j} \quad (12)$$

where: $E_{b_i \to b_j}(t)$ represents the amount of energy traded from the selling microgrid $MG_i$ to the buying microgrid $MG_j$; $\pi_{i \to j}$ is the profit coefficient per unit of energy, defined as follows:

$$\pi_{i \to j} = \begin{cases} 2.5, if E_{b_i}^{final} \in [BT_i, ST_i] \\ 1.2, otherwise \end{cases} \quad (13)$$

In the proposed model, only selling microgrids generate profit, as they are the ones that initiate and control the amount of energy transfers in the community microgrids. Buyers act deterministically according to the energy deficit, and their evaluation is indirectly reflected by the impact that transactions have on their stability.

To encourage solutions that lead to the stabilization of a significant number of microgrids, we introduced a global stability bonus, defined as follows:

$$B = \begin{cases} 0.5 * n, if\ S \geq 0.9n \\ 0.3 * n, if\ S \geq 0.8 * n \\ 0.1 * n, if\ S \geq 0.7 * n \\ 0, otherwise \end{cases} \quad (14)$$

where *n* represents the total number of microgrids in the community.

The total penalty applied to each individual is computed as a linear combination of the following components:

$$P_{total} = w_1 * P_{stability} + w_2 * P_{strategy} + w_3 * P_{noCycles} + w_4 * P_{overhead} \quad (15)$$

where:
- $P_{stability}$ quantifies the deviation of unstable microgrids from the center of their stability interval.
- $P_{strategy}$ is active only for Dove microgrids and penalizes behaviors involving an excessive number of trading partners, violating the conservative nature of the strategy.
- $P_{noCycles}$ reflects battery degradation and is directly proportional to the number of battery charge/discharge cycles in the transaction process.
- $P_{overhead}$ penalizes microgrids that overload distribution lines over a safety threshold.

The coefficients $w_1$, $w_2$, $w_3$, $w_4$ are empirically chosen and represents weights that adjust the relative importance of each component in the final penalty.

The instability penalty, $P_{stability}$, evaluates the deviation of microgrids from their own energy stability range, defined by the $BT_i$ and $ST_i$ thresholds:

$$P_{stability} = \sum_{i=1}^{n} \begin{cases} \frac{|E_{b_i}^{final} - \frac{BT_i + ST_i}{2}|}{E_{b_i}^{max}} , & \text{if } E_{b_i}^{final} < BT_i \vee E_{b_i}^{final} > ST_i \\ 0, & \text{if } BT_i \leq E_{b_i}^{final} \leq ST_i \end{cases} \quad (16)$$

where:
- $E_{b_i}^{final}$ is the energy level stored in the microgrid $MG_i$ after all transactions.
- $BT_i$ and $ST_i$ are the buying and selling thresholds that define the energy stability interval for microgrid $i$.
- $\frac{BT_i + ST_i}{2}$ is the energy level at which the microgrid battery is considered to be in an operationally stable state.
- $E_{b_i}^{max}$ is the maximum capacity of the microgrid battery.

The final energy $E_i^{final}$ is calculated depending on the role played in the transaction as:

$$E_i^{final} = \begin{cases} E_{b_i}^{initial} - \sum_{i \neq j} E_{b_i \rightarrow b_j}(t), & \text{if } MG_i \text{ is seller} \\ E_{b_i}^{initial} + \sum_{i \neq j} E_{b_i \rightarrow b_j}(t), & \text{if } MG_i \text{ is buyer} \end{cases} \quad (17)$$

where $E_{b_i}^{initial}$ is the initial energy level in microgrid battery $i$, before trading, $\sum_{i \neq j} E_{b_i \rightarrow b_j}(t)$ is the total amount of energy sold/bought by microgrid $MG_i$ from/to other microgrids.

The strategy penalty, $P_{strategy}$, is applicable exclusively to microgrids that adopt conservative behaviour (i.e. Dove role) and aims to detect behaviours that contradict the assumed role. In particular, the penalty is activated if a Dove microgrid collaborates with an excessive number of trading partners, which contradicts its prudential behaviour:

$$P_{stratey} = \sum_{i=1}^{n} \begin{cases} \frac{n_i - n_{max}}{n_{max}}, & \text{if } role_i = dove \wedge n_i > n_{max} \\ 0, & \text{otherwise} \end{cases} \quad (18)$$

where: $n_i$ is the number of distinct microgrids to which microgrid $MG_i$ sells energy; $n_{max}$ is the maximum allowed limit of trading partners for a Dove seller microgrid set empirically (in our case is $n_{max}$ =3); $role_i$ is the strategy assigned to seller microgrid

The battery penalty, $P_{noCycles}$, reflects the batteries degradation of the selling microgrids. This penalty is higher in the case of aggressive sales strategies that accelerate the battery degradation. Thus, it functions as a mechanism to discourage aggressive behaviours of sellers that favour immediate profit but damage the energy storage infrastructure.

$$P_{noCycles} = \sum_{i=1}^{n} \frac{noCycles_i^{new} - noCycles_i^{initial}}{noCycles_i^{max}} \quad (19)$$

where:
- $noCycles_i^{initial}$ is the number of discharge cycles of the microgrid battery before trading.
- $noCycles_i^{new}$ is the number of remaining cycles after trading.
- $noCycles_i^{max}$ is the maximum allowed number of charge-discharge cycles of the microgrid battery.

The overhead penalty, $P_{overhead}$ is applies to microgrids that trade large amounts of energy. This penalty reflects the risk associated with overloading the distribution lines and is activated when the total level of energy trading exceeds a predefined tolerance threshold:

$$P_{overhead} = \begin{cases} \sum_{i=1}^{n} \frac{E_{b_i}^{total}(t)}{E_{max-lines}(t)}, & \text{if } E_{b_i}^{total}(t) > \mu * E_{max-lines}(t) \\ 0, & \text{otherwise} \end{cases} \quad (20)$$

where:
- $E_{b_i}^{total}(t) = \sum_{i \neq j} E_{b_i \rightarrow b_j}(t)$ is the total amount of energy traded by microgrid $i$;
- $E_{max-lines}(t)$ represents the maximum allowed energy transfer limit per microgrid.

- $\mu \in [0,1]$ is a tolerance coefficient that defines the activation threshold of the penalty (in our case μ=0.8 and its value was empirically established).

The theoretical maximum value of the fitness function ($F_{max}$) is calculated based on the formula below:

$$F_{max} = \alpha * n * THV * Payoff_{max} + \beta * n + \gamma * B_{max} * n \quad (21)$$

where: *n* is the total number of microgrids; *THV* represents the global transaction threshold; $Payoff_{max}$ is the maximum possible profit that can be achieved in situations where the transactions directly contribute to the energy stabilization of the buyers; $B_{max}$ the maximum value for the stability bonus; $\alpha, \beta, \gamma$ are weight coefficients in the fitness function.

The genetic algorithm ensures the evolution of the population of individuals from one generation to the next generations through three main steps: selection, recombination and mutation. Within the selection process, a combined strategy of elitism and directed random selection is applied. At each generation, a fixed number of individuals with the highest fitness score is kept unchanged, ensuring the preservation of the most promising individuals. Subsequently, to complete the population, pairs of parents are randomly chosen from a restricted subset of the population on which the crossover and mutation operators are applied. This combination offers a balanced compromise between the exploitation and exploitation in the evolution process. For recombination, we used a recombination operator specific to the matrix structure of individuals. A block of consecutive lines is randomly selected from the matrix of one parent and inserted into the corresponding position in the matrix of the other parent. After recombination, each individual $EM \in \mathbb{R}^{nXn}$ is subjected to an element-wise Gaussian mutation operator [2]. The process is controlled by a binary mask matrix $M \in \mathbb{R}^{nXn}$, of the same size as EM, which determines the positions on which the mutation is applied. This element of the mask matrix is defined as follows:

$$M_{ij} = \begin{cases} 1, if\, rand_{ij} < P_{mutation} \wedge EM_{i,j}\, is\, a\, valid\, transaction \\ 0, otherwise \end{cases} \quad (22)$$

where $r_{i,j} \sim U(0,1)$ is a uniformly distributed random variable, $P_{mutation}$ is the current mutation rate and $EM_{i,j}$ is an element of the *EM* matrix.

For each active position in the mask matrix (i.e. $M_{i,j}$=1), a Gaussian variation is applied to the corresponding value in the individual matrix:

$$E_{b_i \rightarrow b_j}(t) = E_{b_i \rightarrow b_j}(t) + \varepsilon_{ij}, if\, m_{ij} = 1, \varepsilon_{ij} \sim N(0, \sigma^2) \quad (23)$$

where $\varepsilon_{ij} \sim N(0, \sigma^2)$ is a Gaussian perturbation with zero mean, and standard deviation σ, modeling local random variations that stimulate diversity in the population.

To respect the physical constraints of the community microgrids and algorithmic constraints, the resulting values are limited to the allowed range of the amount of transferable energy, by applying the following formula:

$$EM_{i,j} \leftarrow \min(\max(E_{b_i \rightarrow b_j}(t), 0), THV) \quad (24)$$

To maintain a balance between exploration and exploitation during evolution, the algorithm uses an adaptive mutation rate, which gradually decreases as generations advance. This rate is defined by the formula:

$$P_{mutation}(g) = \max(0.1 * (1 - \frac{currentGen}{NoGen}), p_{min}) \quad (25)$$

where: $currentGen$ is the index of the current generation, $NoGen$ is the total number of generations, and $p_{min}$ is a lower threshold for the mutation rate (e.g., 0.005).

The algorithm pseudocode is provided in ALGORITHM 1. The algorithm starts by considering the set of seller microgrids and the set of buyer microgrids. In the first step, for each microgrid in the set of sellers the adopted strategy (i.e. hawks or doves) is decided by using a random distribution. Then, the population of individuals is randomly generated. For each individual in the population (i.e. an energy trading matrix), an adjustment is applied, in which the traded energy values are adjusted to respect the stability thresholds of each microgrid (i.e. buying and selling thresholds), as well as the maximum amount of energy that can be transferred in a single interaction. Then the quality of an individual is evaluated with a fitness function composed of following components: the profit obtained by sellers, the level of stabilization achieved in the community microgrids, the bonus granted for reaching the global stability threshold and penalties that reflect the degradation of the microgrids batteries. In each generation, the algorithm keeps a fixed subset of individuals with the best fitness scores, thus forming an elite population that is transferred unchanged to the next generation. The rest of the population is completed by crossover and mutation, applied to randomly selected individuals from a top subset of the current population. This strategy ensures a balance between exploiting the best performing solutions and exploring the search space. Through genetic recombination (crossover), a child is generated that inherit characteristics from both parents. The child is then subjected to a Gaussian mutation which consists of applying a controlled random noise to certain elements of the matrix, with a probability that decreases as the generations progress, favouring exploration in the first generations and exploitation in the last generations. After mutation, the child is subjected to an adjustment step that ensures compliance with the trading constraints. Then, the performance of each child is evaluated by calculating the value of the fitness function. The process is repeated until the maximum number of generations is reached, and at the end the best individual is returned.

---

ALGORITHM 1: Evolutionary Optimization for Energy Trading in Community Microgrid

---

**Inputs**: MGS: set of microgrids; STS: set of selling thresholds; BTS: set of buying thresholds; *n*: microgrids number; noGen: generation number; popSize: population size; THV: maximum tradable energy between two microgrids; σ: standard deviation for Gaussian noise; $p_{min}$: minimum mutation rate; α, β, γ, δ: weighting coefficients for fitness function; $w_1, w_2, w_3, w_4$: weights for penalty components; eliteSize: number of elite individuals preserved in each generation; m: number of individuals to be selected for mating pool

**Output** $indiv_{best}$: best individual

**Begin**
1. Population = INITIALIZE_POPULATION(popSize, n, MGS)
2. bestIndiv = {}
3. for g = 1 to noGen do
4.     Fitness = []
5.     for each indiv in Population do
6.         indiv = ADJUST(indiv, THV, STS, BTS)
7.         fitness = COMPUTE_FITNESS(indiv, α, β, γ, δ, $w_1, w_2, w_3, w_4$)
8.         Fitness= Fitness ⊔ fitness
9.     end for
10.    Elite = SELECT_TOP(Population, Fitness, eliteSize)
11.    Population =Population - Elite
12.    MatingPop= SELECT_TOP(Population, Fitness, m)
12.    NewPop = Elite
13.    while len(NewPop) < popSize do
14.        parent1 = RANDOM_SELECTION(MatingPop)
15.        parent2 = RANDOM_SELECTION(MatingPop)
16.        child = CROSSOVER($parent_1$, $parent_2$)
17.        pmut = COMPUTE_MUTATION_PROBABILITY(g, noGen, $p_{min}$)
18.        child = GAUSSIAN_MUTATION(child, pmut, σ)
19.        child = ADJUST(child, THV, STS, BTS)
20.        fitness = COMPUTE_FITNESS(child, α, β, γ, δ, $w_1, w_2, w_3, w_4$)
21.        NewPop = NewPop ∪ child
22.    end while
23.    Population = NewPop
24.    bestIndiv = UPDATE_BEST(Population, Fitnesses, $indiv_{best}$)
25. end for
26. return $indiv_{best}$
**End**

---

# 4. Performance Evaluation

To validate the proposed algorithm, a simulated dataset consisting of 100 energy microgrids was used. The dataset includes, for each microgrid, the following information: a unique identifier, the associated energy trading strategy (i.e. Dove or Hawk), the level of energy stored in the battery before the start of energy trading, the minimum and maximum energy stability thresholds (BT and ST), the maximum battery capacity and the number of remaining charge/discharge cycles. The BT and ST values differ from one microgrid to another, modelling a realistic scenario in which each battery has its own operational limits. To provide an overview of the initial configuration of the microgrids before trading, Figure 1 shows the battery charge level before the start of the trading process, and Figure 2 illustrates the initial distribution of the number of charge/discharge cycles corresponding to each microgrid.

Figure 1 highlights a considerable variability in the level of energy stored in the batteries of the microgrids, with values ranging from approximately 2 kWh to over 12 kWh. This heterogeneous distribution was generated to reflect possible differences in the initial energy state of the microgrids, before energy trading. By modelling these variations, the behaviour of the proposed algorithm was evaluated in a realistic scenario, in which the microgrids in the community may have different levels of energy stored in the batteries, depending on the local renewable energy production and their own consumption.

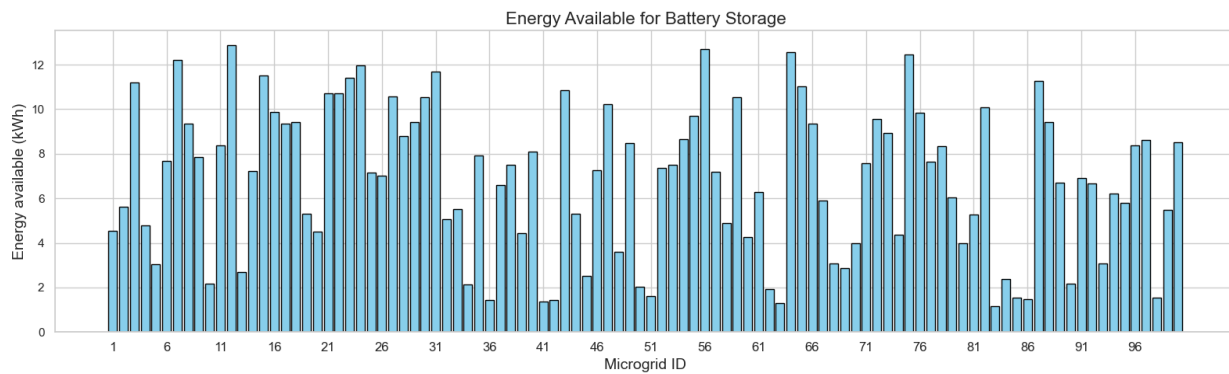

Figure 1: Initial state of charge of microgrids battery

Regarding the distribution of the number of charge/discharge cycles remaining in the microgrid batteries (see Figure 2), before trading it is observed that the values vary between approximately 1000 and over 6000 cycles, reflecting a high degree of diversity in terms of the state of degradation of the batteries.

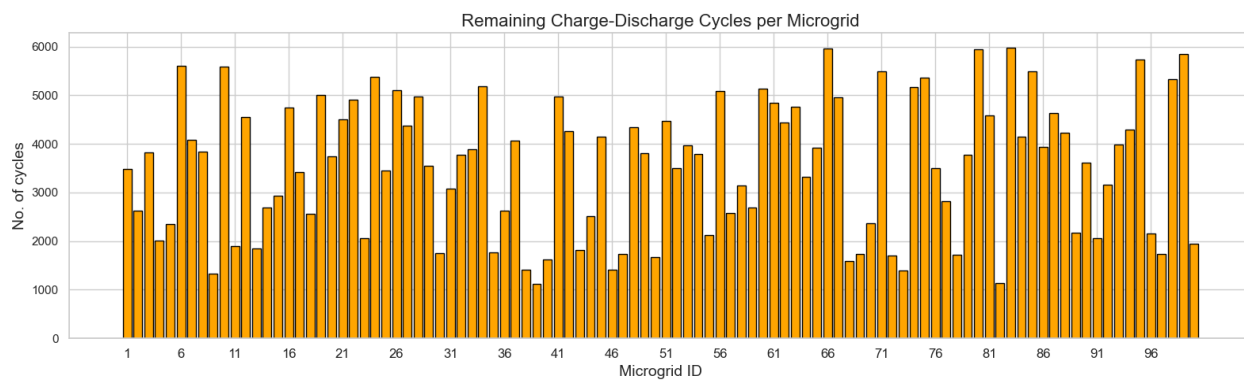

Figure 2: Remaining number of cycles for microgrids batteries

This variability was introduced to test the behaviour of the algorithm in a realistic context, in which microgrids have storage batteries in different phases of their life cycle. The distribution of microgrids according to the role assumed in the transaction and the adopted strategy is established based on the battery energy level and the individual buy (BT) and sell (ST) thresholds. The distribution between roles is relatively balanced, with a slightly higher proportion of microgrids in the buyer role (56%) compared to those in the seller role (44%). Among the selling microgrids, a diversity of trading strategies is observed, reflecting two different behaviours (22% hawk sellers and 22% dove sellers). In the analysis of the performance of the genetic algorithm, we evaluated three aspects: the stability of the microgrid batteries, the total profit obtained by selling microgrids, differentiated according to the strategy adopted, and the energy traded based on the adopted strategies.

Figure 3 illustrates the number of microgrids that are in a stable state versus the number of microgrids that are in an unstable state before and after the trading process. The results show that, after trading, 93 microgrids reach a stable energy state. This performance demonstrates the algorithm ability to identify optimal pairs of selling and buying microgrids between which to conduct energy transactions, thus maximizing the number of microgrids that achieve a state of energy stability at the level of the entire community.

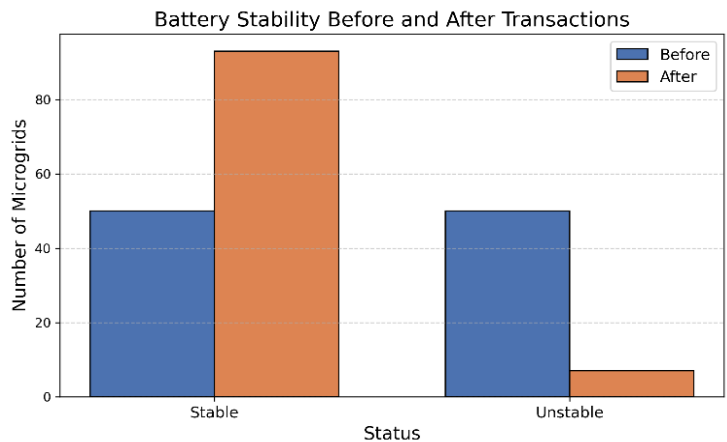

Figure 3: Microgrid battery status before and after transactions

To assess the impact of sales strategies on the economic performance of selling microgrids, we analyzed the total profit generated from transactions with buying microgrids, differentiated according to the strategy adopted by the sellers (i.e. Hawk or Dove). Figure 5 illustrates the total profit generated by each buyer microgrid for sellers, separated according to the strategy adopted by the sellers. The Ox axis indicates the buying microgrids, labeled with their ID and current strategy, and the Oy axis represents the total profit generated from trading interactions. Each vertical bar represents the sum of the profits obtained by individual sellers from transactions with that buyer. The colors code represents the sellers' strategy: green for Dove and purple for Hawk. The results show that most buyers tend to generate higher profits for Hawk sellers. This suggests that aggressive sales strategies can lead to higher economic gains, even if the risk of battery damage is higher. However, there are also some exceptions that show cases were buying microgrids have generated comparable or even higher profits for Dove sellers, indicating that the profit is not dictated solely by the aggressive strategy, but also by the context of the interaction. In addition, a clear difference in variability is observed between the two selling strategies. Dove sellers obtain profit that does not fluctuate, while Hawk sellers record much more fluctuating profits. This difference suggests that the Dove strategy offers a more stable trading framework, while the Hawk strategy involves higher profits.

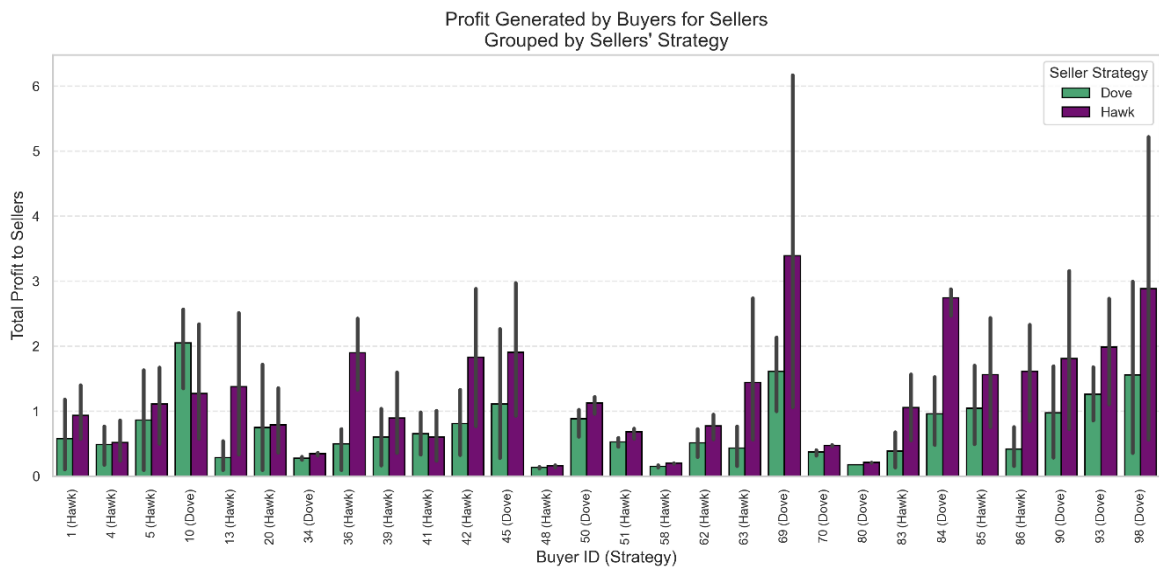

Figure 4: Profit generated by buyers for sellers

Figure 6 shows the total quantity of energy traded by each microgrid within the community, providing an overview of the degree of individual involvement in the energy trading processes. The graph includes both selling and buying microgrids. However, only sellers adopt selling strategies, and these strategies influence the trading process. It is observed that selling microgrids that adopt the Hawk strategy are generally involved in trading larger volumes of energy, reflecting an approach oriented towards maximizing profit, even with higher risks. In contrast, Dove sellers trade more moderate volumes, acting cautiously to protect battery degradation. For buying microgrids, which do not adopt individual strategies, the amount traded exclusively reflects the energy requirement to reach equilibrium. Thus, high values of traded energy for certain buyers are associated with large deficits. This distribution of transaction involvement highlights the ability of the genetic algorithm to generate an optimal configuration of energy trading between microgrids, in which trading decisions are made based on the energy state of each microgrid and the role that each microgrid plays in the trading process.

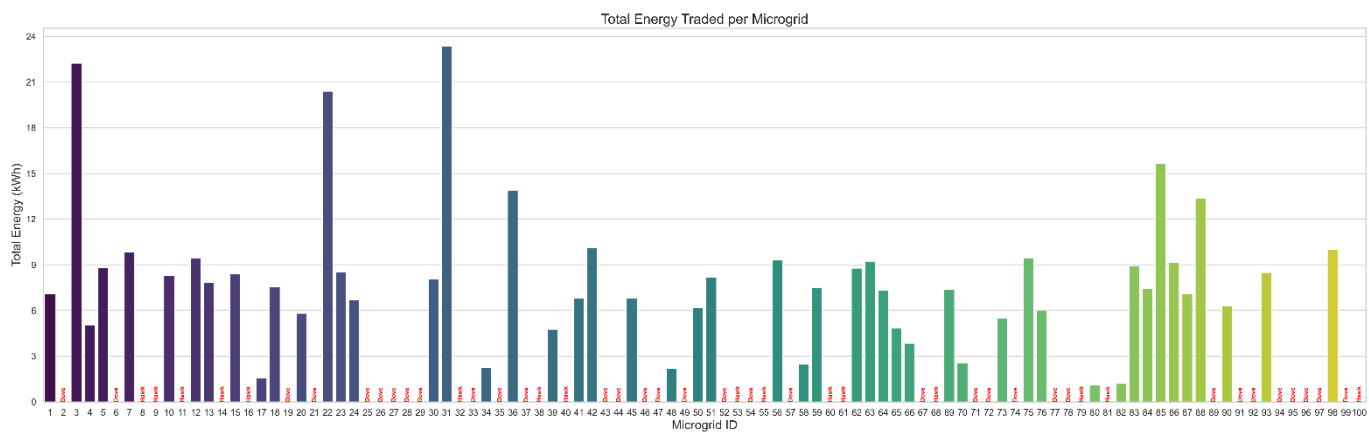

Figure 5: Total Energy traded per microgrid

# 5. Discussion

To evaluate the performance of the genetic algorithm, we conducted a detailed analysis of multiple configurations of the GA's adjustable parameters, aiming to identify the optimal configuration in terms of fitness value, population diversity, and execution time. Once the optimal configuration was established, we investigated the balance between exploration and exploitation, the stability of the algorithm between successive runs for the same initial population, as well as its variability under different initializations and random seeds. To identify an efficient configuration of the genetic algorithm parameters, we adopted a greedy search strategy, in which several combinations of parameters were tested. The parameters and the ranges in which they were varied are the following: popSize ∈ [60,100], noGenerations ∈ [400,600] mutation rates∈ [0.005, 0.007, 0.01], and elite size ∈ [10-14]. In parallel, the seed parameter was evaluated separately through a series of preliminary experiments, with the aim of identifying a value that would ensure both high performance and stability in the behaviour of the algorithm. Since seed determines the initialization of the random generator and directly influences the initial population, several values were tested. Based on these tests, the value 120 was selected for the final experiments. For each configuration evaluated in the greedy search, the following were analysed: the maximum value of the fitness function, the total execution time and the number of microgrids reaching an energy stability state. Following this process, the five best configurations were selected (see Table 1), which were analysed in detail from the perspective of the fitness evolution over generations, and the diversity evolution in the population.

Table 1: The best configurations of adjustable parameters for GA algorithm identified with Meta – GA

| ConfigID | GeneNo | popSize | EliteSize | MinMutation | Fitness | No MG | Time (sec) |
|---|---|---|---|---|---|---|---|
| 1 | 500 | 80 | 13 | 0.005 | 0.7533 | 93 | 203.38 |
| 2 | 500 | 80 | 12 | 0.007 | 0.73 | 91 | 201.9 |
| 3 | 500 | 80 | 12 | 0.01 | 0.7369 | 91 | 191.5 |
| 4 | 400 | 80 | 14 | 0.005 | 0.7505 | 90 | 163.6 |
| 5 | 400 | 80 | 14 | 0.007 | 0.7481 | 90 | 164.83 |

Figure 7 shows the evolution of the maximum fitness value for the top five tested configurations. The G500_P80_E13_M0.005 configuration is among those that achieve the best fitness. Although the G500_P80_E12_M0.01 configuration achieves a slightly higher fitness value, Figure 8 highlights a slower decrease in diversity in the case of the G500_P80_E13_M0.005 configuration, suggesting a better balance between exploration and exploitation. Based on this analysis, the G500_P80_E13_M0.005 configuration was selected as optimal, offering a good compromise between the quality of the solution obtained and the maintenance of population diversity. The fitness scores obtained, ranging from 0.73 to 0.75 for the best configurations, reflect a good performance of the genetic algorithm. These values correspond to scenarios in which between 90 and 93 out of 100 microgrids reach the state of energy stability following the trading process. It should be noted that the theoretical maximum value of 1 for fitness is associated with an ideal case, in which all microgrids simultaneously reach stability, without penalties related to strategy, instability or transaction costs. Thus, the subunit values recorded are explained by the penalties applied to suboptimal solutions.

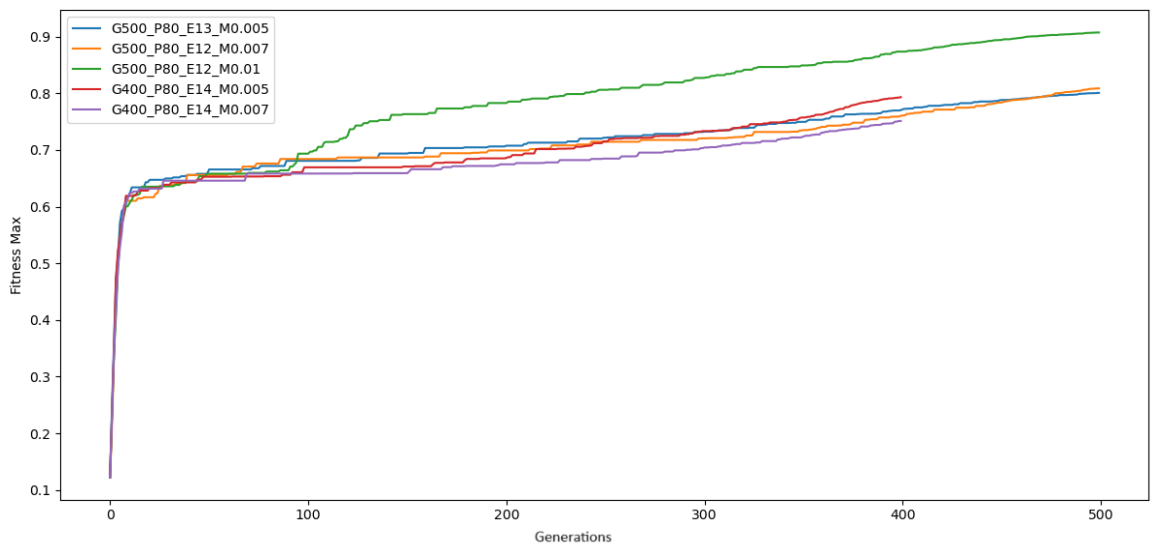

Figure 6: Fitness evolution over generations for the best configurations of adjustable parameters

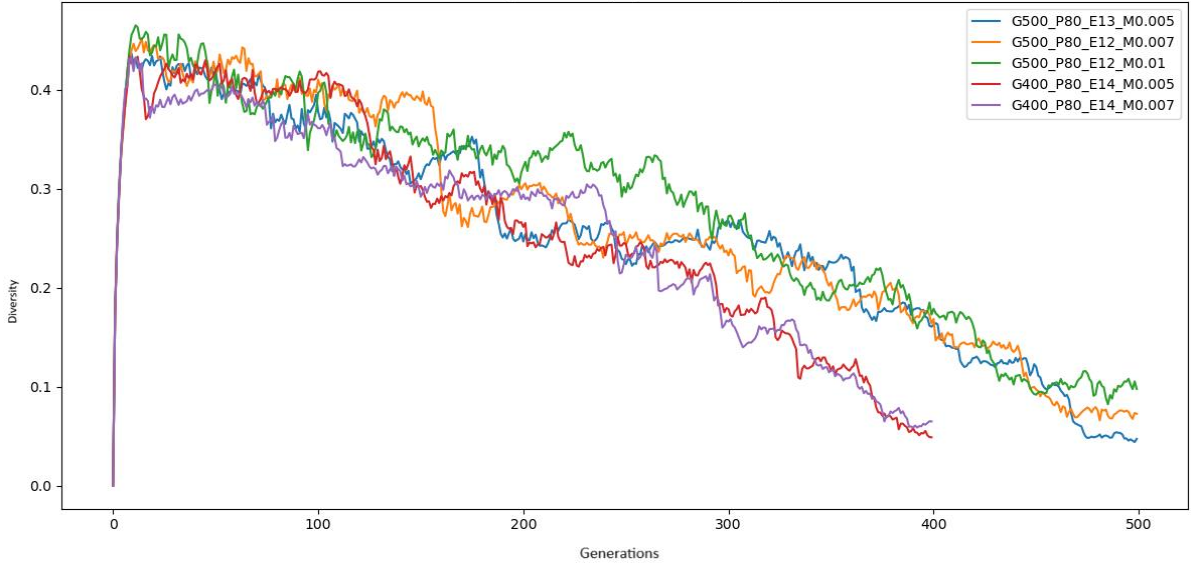

Figure 7: Diversity evolution over generations for the best configurations of adjustable parameters

For the identified optimal configuration of adjustable parameters, we conducted an additional analysis on the balance between exploration and exploitation, as well as the stability of the genetic algorithm. To quantify the balance between exploration and exploitation throughout the algorithm evolution, the percentage values associated with each component in each generation were calculated, using the following formulas [100]:

$$exploration\% = \frac{diversity}{diversity_{max}} * 100 \quad (30)$$

$$exploitation\% = \frac{|diversity - diversity_{max}|}{diversity_{max}} * 100 \quad (31)$$

where *diversity* represents the value of the population diversity in each generation, measured as the average distance between individuals, and *diversity*$_{max}$ is the maximum value of diversity recorded over the entire run of the algorithm.

Figure 9 illustrates the percentage of exploration vs. exploitation over 500 generations. In the initial stages, the algorithm exploratory more regions of the search space, reflected by a high level of population diversity. As the generations advance, a gradual transition towards exploitation is observed, indicating a progressive focus on refining already discovered solutions. This behaviour indicates that the algorithm maintain a good balance between the exploration and exploitation, an essential aspect for preventing premature stagnation and for converging towards high quality solutions.

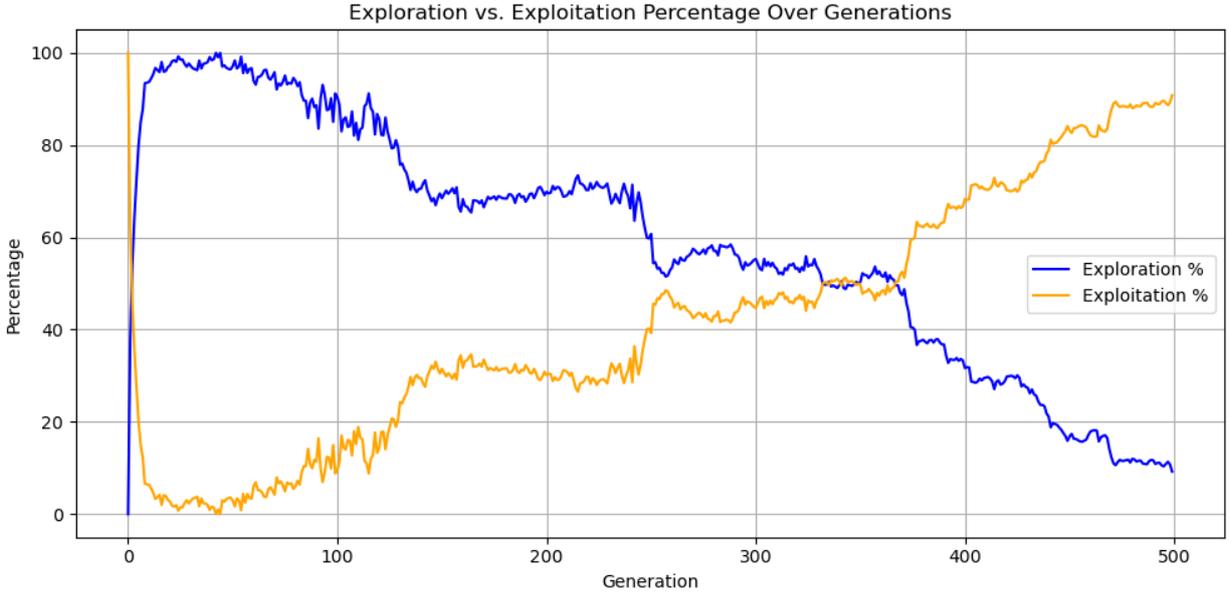

Figure 8: Exploration vs exploitation when running GA for optimal configuration of adjustable parameters

To evaluate the stability of the genetic algorithm (GA), we conducted an experiment designed to test its behaviour under various conditions. The experiment consisted of running the algorithm multiple times, using the same initial population for each execution. Figure 10 compares the final fitness values obtained from five independent runs of the algorithm, using the optimal parameter configuration.

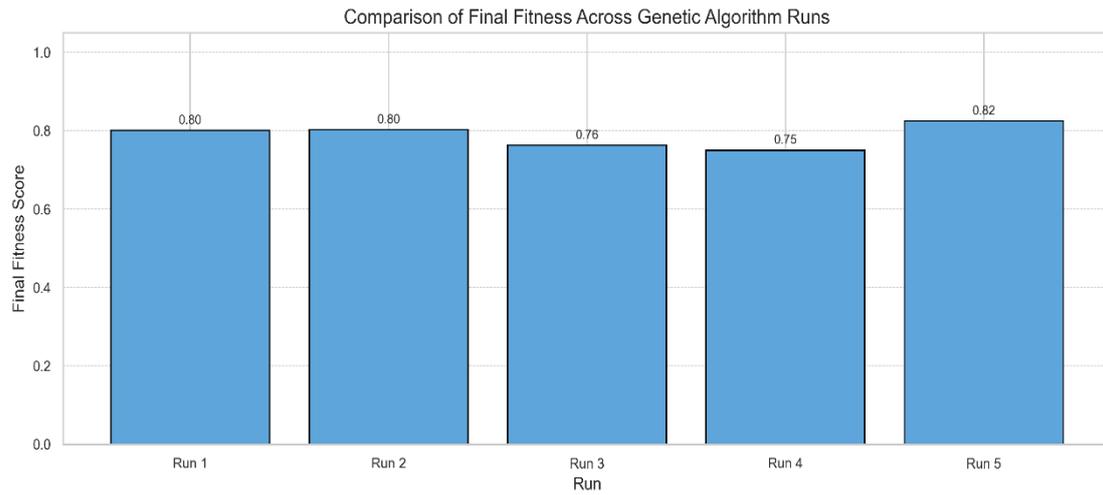

Figure 9: Comparison of fitness values for different running of GA when considering the same initial population

A low variability between runs is observed, with values ranging from 0.75 to 0.82, which highlights the stability of the algorithm in the case of random variations induced by the population initialization. The results indicate that the fitness obtained from running the algorithm is not just an isolated result but reflects its ability to generate reproducible results under similar execution conditions.

# 6. Conclusion

This paper proposed a decentralized energy trading model between microgrids, in which each microgrid operates autonomously, in the absence of a central coordinator. The model combines evolutionary game theory with a genetic algorithm to optimize energy trading between selling and buying microgrids, with the aim of ensuring energy stability at both the individual and community level. In this approach, microgrid sellers adopt Hawk or Dove trading strategies, depending on the level of energy stored in the battery. Microgrid buyers who have an energy deficit do not have an associated strategy but only aim to restore energy balance. In the genetic algorithm, an individual is represented by an energy trading matrix, and the optimization is performed based on a multi-criteria fitness function that evaluates sellers' profit, community microgrids stability, microgrids battery degradation and the level of infrastructure load. The method was validated through a simulated scenario with 100 microgrids, each with batteries with specific properties. The results obtained show that 95 of the microgrids reached a stable energy state, confirming the efficiency of the algorithm in achieving energy balance at the community level. Unlike centralized methods or those that do not consider individual trading strategies, the algorithm explicitly models the behaviors of microgrids and allows energy transactions to be made completely decentralized, without the intervention of a central coordinator. As a future direction, we aim to introduce an adaptive learning mechanism at the level of each microgrid, which would allow updating trading strategies based on interaction history and consumption and production forecasts. This adaptive capacity could significantly improve the algorithm's performance in dynamic and unpredictable trading environments.